\documentclass[conference]{IEEEtran}

\usepackage[english]{babel}
 
\usepackage{multicol}
\usepackage{helvet}
\usepackage[utf8]{inputenc}
\usepackage{graphicx}
\usepackage{geometry}
\usepackage{hyperref}
\usepackage{amsmath}
\usepackage{float}
\graphicspath{ {images/} }

\hypersetup{
    colorlinks=true,
    linkcolor=blue,
    filecolor=magenta,      
    urlcolor=cyan,
}

\begin{document}
\title{Enhanced Image Classification With Data Augmentation Using Position Coordinates}

\author{\IEEEauthorblockN{Avinash Kori}
\IEEEauthorblockA{avinashgkori@smail.iitm.ac.in}
\and
\IEEEauthorblockN{Ganapathy Krishnamurthi}
\IEEEauthorblockA{gankrish@iitm.ac.in}
\and
\IEEEauthorblockN{Balaji Srinivasan}
\IEEEauthorblockA{sbalaji@iitm.ac.in}}
\maketitle
\begin{abstract}


In this paper we propose the use of image pixel position coordinate system to improve image classification accuracy in various applications. Specifically, we hypothesize that the use of pixel coordinates will lead to (a) Resolution invariant performance. Here, by resolution we mean the spacing between the pixels rather than the size of the image matrix. (b) Overall improvement in classification accuracy in comparison with network models trained without local pixel coordinates. This is due to position coordinates enabling the network to learn relationship between parts of objects, mimicking the human vision system. We demonstrate our hypothesis using empirical results and intuitive explanations of the feature maps learnt by deep neural networks. Specifically, our approach showed improvements in MNIST digit classification and beats state of the results on the SVHN database. We also show that the performance of our networks is unaffected despite training the same using blurred images of the MNIST database and predicting on the high resolution database. 
\end{abstract}
\IEEEpeerreviewmaketitle 
\section{Introduction}

Image classification using images acquired with different pixel spacing or scales is a difficult problem and is in general solved using very deep neural networks trained with a large amount of data. However, in many problems of interest such large data is not readily available and alternate approaches are desirable. Typically, CNNs are trained using a multi-resolution approach wherein images down sampled or blurred to various resolutions, these whole lot of images with different resolutions are used to train network and their predictions are used for final classification. 
However, this leads to large networks and a cumbersome training process. 

The capability to classify images at various resolutions is, however, achieved almost seamlessly by the human vision system. One model of human perception is based on sets of neurons representing hierarchical representations of objects starting with basic shapes. For instance, certain cells in the human retina are sensitive to horizontal and vertical lines which can be used to compose basic shapes\cite{RetinalVision}. CNNs mimic this model by using layers of filter to construct  hiearchical representations and the presence of a local coordinate system  in the input would definitely help learn these representations better by learning relationships between parts of objects. One key difference between typical CNNs and the human vision system is that position information is more immediately present in the latter due to  the geometric arrangement of neurons; In the case of usual neural networks position information can only be learnt (and that too, implicitly) via training as, typically, any image is unrolled into a single column vector before training. Similarly, the level of explicit coordinate position encoded into even CNNs is limited by the size of the filter. 

In this article, we explore the advantages of mimicing the geometric feature of the human vision system by explicitly encoding a normalized coordinate system into the CNN. We hypothesize that introducing a local pixel coordinate system on the images, i.e. a XY cartesian coordinate system and using it as one of the input channels would enable the network to learn the relationships between parts of objects. We expect, in turn, that this recognition would enhance performance in both image classification as well in multi-resolution problems. In a simple sense, one would expect the network to learn Euclidean distances between significant features in an object. As a result, an improvement in classification accuracy can be expected. One would also expect an invariance to pixel spacing, as recognition should be based on relative distances rather than on the actual resolution of the images. 

In our work we test the use of local pixel coordinates using commonly used digit datasets like MNIST\cite{MNISTDatabase}, SVHN\cite{SVHNDatabase}( MNIST is a benchmark dataset for an images of segmented handwritten digits, each with 28x28 pixels. It includes 60,000 training examples and 10,000 testing examples, Figure \ref{mnist_sample}. The SVHN dataset is a real life application oriented dataset for digits which includes images with 32x32 pixels, with 73257 digits for training and 26032 digits for testing, Figure \ref{svhn_sample}),
and show empirical evidence in terms of improvement in classification accuracy over regular approaches. In the case of SVHN, we demonstrate the capability to beat state of the art results using position coordinates as input to a deep neural network. We also examine feature maps for less complex networks like the one used for digit recognition and we demonstrate how these networks combine coordinate data and shape information. 
\section{Methodology}

\begin{figure}[H]
\centering
\includegraphics[width=9cm]{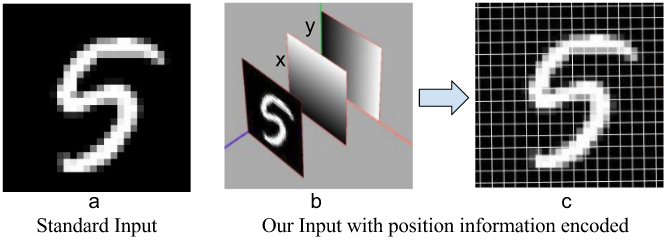}
\caption{a. Standard input image used |
b. Shows the layers of position information with an input image | 
c. Our input image encoded with position information
}
\label{Fig 1: input_image}
\end{figure}

We discuss below our approach to encoding position information and creating tests for multi-resolution efficacy.

a) Encoding Position Information: The CNN is fed with a 3 channel input, the first channel comprises of the actual input image (gray-scale image). The remaining two channels are for encoding spatial position along X and Y dimension with respect to Cartesian coordinate frame of reference(auxillary frames). These position encoded image frames are spatial coordinate grids normalized between 0 and 1.  Figure: \ref{Fig 1: input_image} . illustrates this. 

b) Validating multi-resolution behavior : In this analysis we trained the network with images of one specific resolution and in testing phase, we used images of entirely different resolution. 
In this study of multi-resolution behavior, the previously shown method is followed however the coordinate grid is also resized to the same resolution as that of an image. In many applications involving images, we train the model using images of a fixed resolution; but for most of the practical and real life applications there will  always be  varying resolutions, frame of reference
which need to be detected, due to which most of the models fails. Since the downsampled coordinate information is obviously image independent, position encoding is a potentially promising approach to tackle such scenarios. In the next subsection we discuss how the results were improved just by encoding position information with input image.\\
 
\section {Experiments and Results}

Our hypothesis was that encoding position information explicitly would improve both image classification as well as multi-resolution performance. We tested our hypothesis through the relative performance of our network in comparison to conventional networks on the following tasks:
\begin{enumerate}
\item Classification accuracy on the MNIST dataset
\item Classification accuracy on the SVHN dataset
\item Digit translation test
\item Multi-resolution test
\end{enumerate}

\subsection{Test On MNIST Dataset}

\begin{figure}[H]
\centering
\includegraphics[width=3cm]{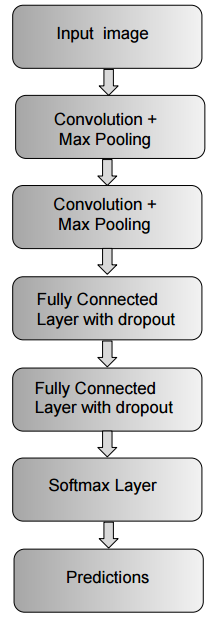}
\caption{MNIST network | architecture details: convolutional layer : 5x5x32| pooling1: 2x2|convolutional layer : 5x5x64| pooling2: 2x2  }
\label{architecture_mnist}
\end{figure}

\begin{figure}[H]
\centering
\includegraphics[width=5cm]{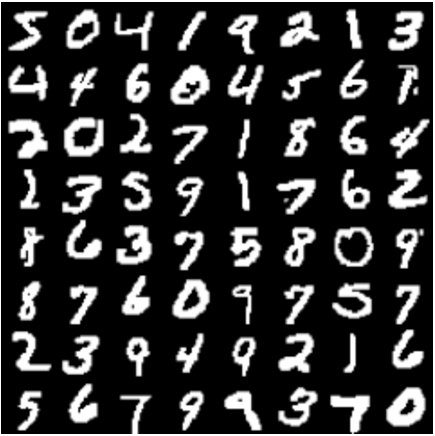}
\caption{examples of MNIST digits}
\label{mnist_sample}
\end{figure}

\begin{figure}[H]
\centering
\includegraphics[width=5cm]{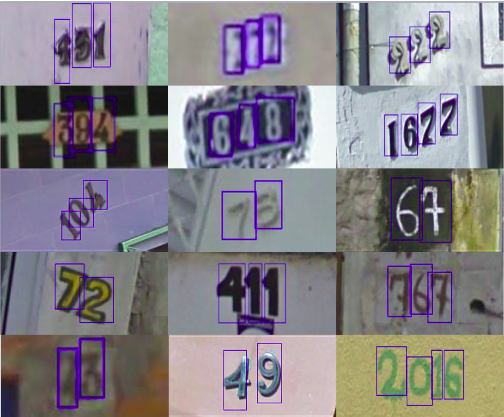}
\caption{examples of SVHN digits}
\label{svhn_sample}
\end{figure}

A schematic of the network architecture followed for MNIST digit classification is shown in Figure \ref{architecture_mnist}. In this network we use two sets convolutional layers with maxpooling layer followed by two fully connected layers with softmax layer to obtain the  posterior probability of an input digit.
On MNIST handwritten digits dataset, with 45,000 images for training, 5,000 for validation and 10,000 images for testing, with our auxiliary input
(XY-Position matrix) channels, We were able to achieve 99.84\% accuracy. This was about 0.2\% greater than maximum accuracy achieved using a conventional network without using position information\cite{MNISTResult}. In other words, the error rate was reduced by approximately a factor of 2 due to the use of position information.


\subsection{Test On SVHN Dataset}

In the SVHN classification task, we set an upper limit for number of digits in the image as 5, and remove all other images with more than 5 digits from dataset and didn't use them either in training or for testing.
We use VGG16 (16 layered deep convolutional network which is used for image classification tasks) convolutional architecture \cite{simonyan2014very} for feature extraction and we use attention based modeling \cite{kim2017structured} for predicting existing digits in an image. The input image is appended with XY position information and all the digits are predicted simultaneously. The entire model was trained end-to-end without data augmentation. In attention mechanism is basically weighted superposition of higher level features extracted by VGG16 convolutional neural network.

\begin{figure}[H]
\centering
\includegraphics[width=6cm]{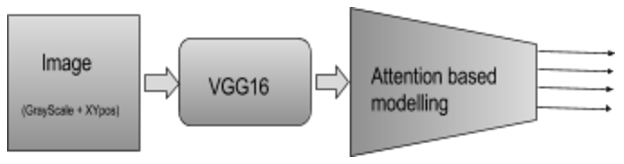}
\caption{Network architecture followed for SVHN classification}
\label{svhn_architecture}
\end{figure}


Typically, in CNNs\cite{ConvolutionalLayer} used for digit recognition tasks, features learnt by network after a few convolutional layers would  correspond to entire digit itself. However, the learnt features could also be combination of multiple digits. The attention model used in our network has learnable parameters which helps in providing specific weightage to feature maps in predicting final output. Our results indicate that the use of attention model mechanism has a positive impact on the classification accuracy. 

We were able to reduce the the digit recognition MSE by 1.2\%, by including the XY position channel information. On a real life dataset like SVHN, with auxiliary (XY-position matrix) channels, we achieved around 95.44\%(average of all 5 digits) accuracy with 1.83\% classification error. Whereas the current state of the art result is 94.6\%(average of all 5 digits) accuracy with 1.64\% classification error\cite{sermanet2012convolutional}. 

%
\subsection{Resolution Change Test}
Our next endeavor was to show that using XY position information makes the network resolution invariant. Networks were trained with MNIST digits whose resolution was artificially degraded by blurring and tested with digits at the original resolution. The classification results indicate that implementing position matrix aided in achieving exceedingly good outputs (improvement by 2-3\%) despite mismatch in training and testing resolution. 

\begin{figure}[H]
\centering
\includegraphics[width=6cm]{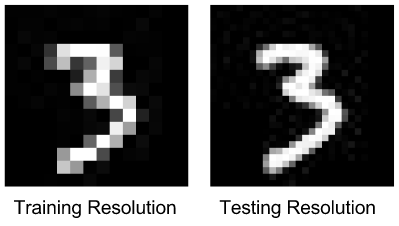}
\caption{Variation in resolution among training and testing images }
\label{res_test}
\end{figure}
Training images were obtained by down sampling the original 28x28 image to 14x14 and again up sampling it to the original resolution. Basically, the network is trained on low resolution images and tested on high resolution images (original data). In this case, adding XY position information, in addition to gray channel, increases digit classification accuracy by 2-3\%.  With this kind of data our network with encoded position information gives an accuracy about 98.26\% while the standard network gives an accuracy of about 96.32\%. Once again, the error rate was reduced by approximately a factor of 2 due to the use of position information.

%

\subsection{Digit Translation Test}
In this test we show how translation invariance can be obtained by using by employing XY position information with an image. (which is equivalent to labeling pixels in an image) our model could predict digits correctly, regardless of finite translations. 
This would not be possible to accomplish merely with a standard input or even with max pooling (without position information)
Some of these results are shown in the figure below.

\begin{figure}[H]
\centering
\includegraphics[width=6.5cm]{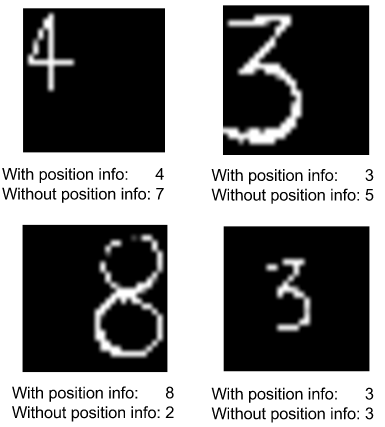}
\caption{Example images which are used in translation test}
\label{trans_test}
\end{figure}
The results obtained so far on MNIST, SVHN databases and the translational invariance experiment indicate that the XY coordinate information improves prediction accuracy. We assert that using XY inputs could also help to reduce the number  of parameters required for achieving a certain level of accuracy. The following numerical experiments demonstrate that using XY position input can help reduce the number of network weights.
%

\section{Discussion and Results}
\subsection{What does this position matrix learn?}
\begin{figure}[H]
\centering
\includegraphics[width=8cm]{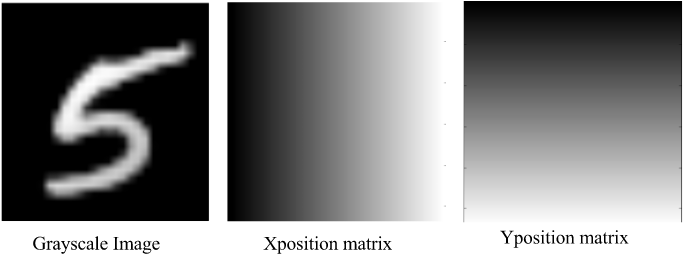}
\caption{Normalized input given to Variational Auto Encoder}
\label{vae_input}
\end{figure}

\begin{figure}[H]
\centering
\includegraphics[width=8cm]{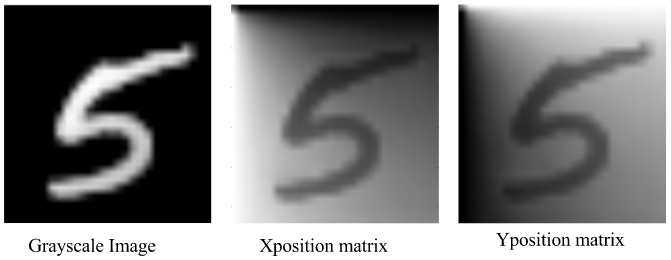}
\caption{Variational Auto Encoder Output (After Reconstruction) }
\label{vae_output}
\end{figure}

To see what exactly could be learnt from the  position matrix, we implemented convolutional variational auto-encoders(\cite{makhzani2015adversarial} \& \cite{doersch2016tutorial}). These auxillary layer (XY position matrix) helps
in removing useless information(noise) for any given data. Figure \ref{vae_input} \& figure \ref{vae_output} shows input and corresponding output for convolutional variational auto-encoders experiment. The reconstructed outputs are revealing as the reconstruction of the position matrix contains the digits superimposed on it.\\
%

\begin{figure}[H]
\centering
\includegraphics[width=5cm]{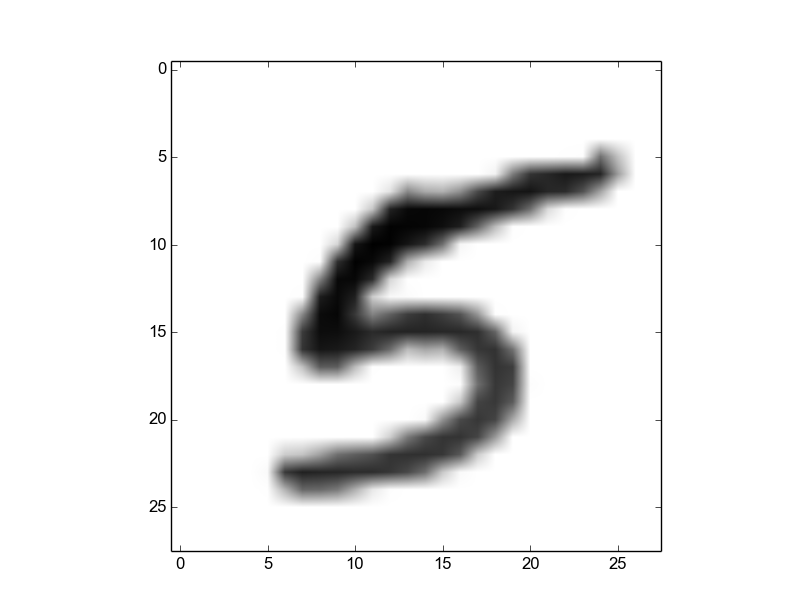}
\caption{Joint position matrix(Sum of reconstructed x and y position matrix)}
\label{vae_joint}
\end{figure}
As the input and output images are normalized to unity, the black pixel values correspond to lower probability values and white pixel correspond to higher probability information. As the XY position matrix was auxiliary input to the network there are no specific features which helps in reconstruction of these matrices. But at the same time, as mentioned earlier we can see that the reconstructed position matrices are related to the corresponding input digit image. If we consider the sum of the reconstructed position matrices as the joint probabilities of X and Y matrix (As all image pixels are bounded between 0 and 1, and pixel probability distribution is linear(uniform to certain extent))
then the input image is clearly reconstructed. 
As a result of which, we conclude that the position matrix helps in learning joint probability distribution of the pixels constituting the digits.

\subsection{What is the significance of using XY Position Information?}
\begin{figure}[H]
\centering
\includegraphics[width=8cm]{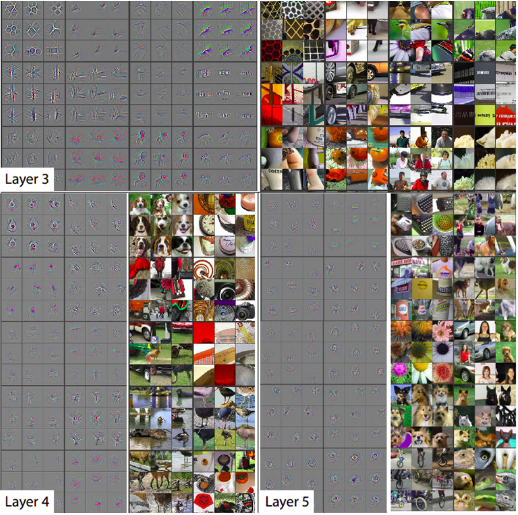}
\caption{The layer wise loss of edge information can be clearly seen in the above image.}
\label{conv_info_loss}
\end{figure}
In standard convolution operation of any image, 
only the local (bound to kernel size) relative information is captured. It takes many number of layers to capture the global relative pixel information with some loss of edge information at each convolutional step.\\
But by adding an XY position layer as a channel information, capturing global pixel relative information becomes easier, and the information is procured at a very early stage. As the position information is encoded within the input image channel, the network learns some sort of features or patterns and their relative positions which makes it possible to capture global information at the very initial layers.\\
Mathematically, for any image, it will take about 
\begin{equation}[\frac{n-s}{k}]\end{equation}
steps of convolution to capture whole global information(without pooling). 
[where, n is the image dimension(assuming image is square),
s is the stride size,and k is the kernel size]

Mathematically Convolution is defined as:
\begin{equation}
Y(i,j) = \sum_{i}\sum_{j}\sum_{ki}\sum_{kj} I(i+ki,j+kj)*K(ki,kj), 
\end{equation} 
where I is the image and K is the kernel 
Convolution on an image with XY position stacked is 
\begin{gather*}
Y(i,j) = \sum_{i}\sum_{j}\sum_{ki}\sum_{kj}I(i+ki,j+kj,x,y)*K(ki,kj),\\
\Rightarrow \sum_{i}\sum_{j}\sum_{ki}\sum_{kj}I(i+ki, j+kj)*K(ki,kj)+\\
				\delta (x,y,i+ki,j+kj)\\
\end{gather*} 
 
The term 
\begin{equation}
 \delta(x,y,i+ki,j+kj)
\end{equation}
saves the encoded XY position information, technically saving pixel-to-pixel relative information which is preserved after successive convolutions, and can enable the network to learn the relative position between features. 
Another observation was that the feature maps learnt by networks using XY position information was less redundant as compared to feature maps of standard image input.

\begin{figure}[H]
\centering
\includegraphics[width=8cm]{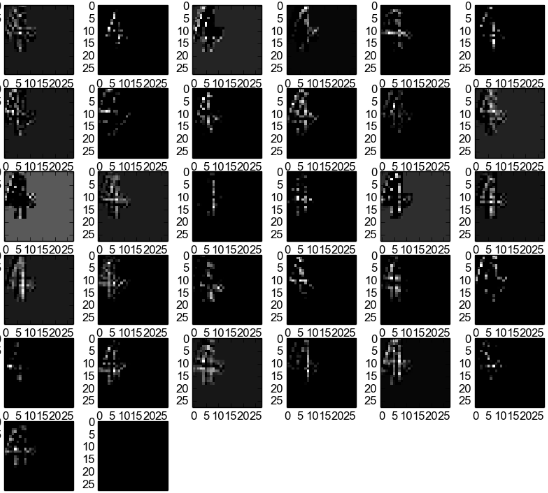}
\caption{Without position information}
\label{features_without_pos}
\end{figure}

\begin{figure}[H]
\centering
\includegraphics[width=8cm]{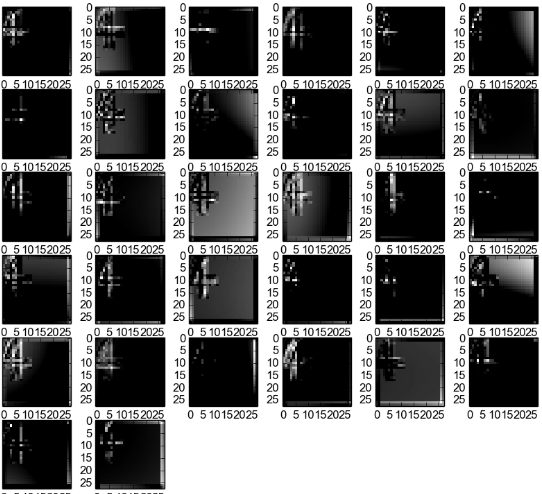}
\caption{With position information}
\label{features_with_pos}
\end{figure}

The figure \ref{features_without_pos} and \ref{features_with_pos} shows  feature maps obtained after the first convolutional layer. It can be observed that quite a few of the feature maps on the top image(without position info.) are fully blank (corresponding to redundant information), but our network using position information was still able to extract  some unique features, as is evident from some of the feature maps shown in the bottom row(with position info.). Using position information seems to help the network learn features that are normally learnt at deeper layers. Consequently using position helps to reduce the depth and consequently the number of weights making the network computationally less expensive.

\section{Conclusions and Future work}
We have demonstrated that implementing position coordinates as an auxillary input results in enhanced accuracy as well improved mutliresolution performance. Our results consistently provide an improvement over state of the art results on publicly available classification datasets.
We aim to extend this study from gray images to colored images and test the result on CIFAR-10 and STL-10 datasets. We also expect that improved performance could be obtained by integrating the position information into a complex number i.e. by replacing XY position information with Z (i.e X+iY). We expect that implementing position information via the complex domain will helps in reducing the number of parameters as well as number of layers.
\\

\bibliography{references} 
\bibliographystyle{ieeetr}




%
\end{document}